# Memory Retrieval in the B-Matrix Neural Network

Prerana Laddha

**Abstract**

This paper is an extension to the memory retrieval procedure of the B-Matrix approach [6],[17] to neural network learning. The B-Matrix is a part of the interconnection matrix generated from the Hebbian neural network, and in memory retrieval, the B-matrix is clamped with a small fragment of the memory. The fragment gradually enlarges by means of feedback, until the entire vector is obtained. In this paper, we propose the use of delta learning to enhance the retrieval rate of the stored memories.

## I Introduction

The brain consists of interconnection of neurons. In an effort to create computing structures that are as efficient as the brain at cognitive tasks, interconnected artificial neurons are used in research in cognitive science and artificial intelligence [1]-[12]. Specifically, research has been done on anomalous abilities [1], [2] and generalization and learning of memories [3]-[16].

The connections between neurons store memories. In mathematical terms, if an $i^{th}$ neuron sends a signal to the $j^{th}$ neuron and the connection weight from neuron $i$ to neuron $j$ have value $w_{ji}$, then the total activity received by the $j^{th}$ neuron is given as

$$A = \sum_i w_{ji} u_i$$

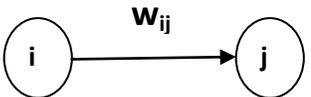

where $u_i$ is the activity of the $i^{th}$ neuron, being 1 if it is active and 0 if it is inactive. The connection weight value $w_{ji}$ determines the magnitude of the signal reaching the $j^{th}$ neuron from the $i^{th}$ neuron [3]. This elementary notion is used in the training and construction of models of both biological neural networks and artificial neural networks.

Artificial neural networks, in an abstract way, are adaptive networks with associative memories. They are associative because when an input is fed into the network a corresponding output or identification is returned and they are adaptive as we can train the network using learning algorithms to produce the required outputs. The input to an artificial neural network is an $N$-dimensional input vector and has one single output signal. The output signal is a non-linear function of the input vector and the weight vector [12].

This paper is an extension of Kak's B-Matrix approach [6],[17], which explains the spreading activity from one neuron to another neuron based on the adjacency of the neurons. The indexing of neurons in this approach is defined by the proximity matrix, defined by the distances in between the neurons in that network [17]. In this paper we propose a method to increase the memory retrieval rate in the B-matrix and this is achieved by applying a learning algorithm to the network which iteratively makes small changes in the synaptic weights of the network resulting in more number of active neurons.

In the rest of the paper, we give a brief literature review in the second section. In the third section we propose our learning algorithm and infer other properties from the resulting network. In the fourth section we give a graphical representation of the results for the B-Matrix and the last section concludes the paper.

## II Literature review

**B-Matrix Approach**

In Kak's B-Matrix, activity spreads locally [15] guided by the adjacency of neurons in the network [7], [17]. The Interconnection matrix is Hebbian:

$$T = \sum_{i=0}^{n} x^{(i)} \, x^{t(i)}$$

The number of memory vectors is *n* and $x^{(i)}$ represents the memory in column vectors that are stored in the network and $x^{t(i)}$ is its transpose. Once the memories are stored in the T-matrix, the B-matrix is calculated in order to initiate the retrieval process of the memory vectors. The B-matrix is the lower triangular matrix of the T-matrix, given as [7]:

$$T = B + B^t$$

A fragment of the memory vector say *1* or *-1* is clamped onto the B-matrix which is structured according to the indexing of a specific neuron in the network. The resultant fragment is fed back into the network to generate a relatively larger fragment and this process is continued until a complete memory vector is obtained. The updating process proceeds as follows:

$$f^i = sgn(Bf^{(i-1)})$$

Where $f^i$ is the $i^{th}$ iteration of the generator model. The proximity matrix of a network plays a substantial role in the B-Matrix approach.

**Delta Rule**

We extend the previous work of Lingashetty [19],[20] by applying the delta rule on the B-matrix method of vector retrieval.

**III Proposed Work**

The fundamental idea in neural networks is that the weights control the response of the output at each stage in the network. The weight values in the neural network have to be modified such that maximum number of memories can be retrieved. *Learning* in a network is a repetitive incremental process by which the system produces desired outputs in response to a specified input. Biologically, synaptic plasticity supports the idea at the basis of neural network learning [21].

**Algorithm**

In the B-matrix approach, we observed how a memory vector can be retrieved by triggering a neuron with a small input fragment. We would like to make small changes to the weights of the network if that results in increased retrieval rate. Consider a network represented as a quadruple, *T = {n, E, W, r,}*, where *n* is the number of nodes, *E* is the number of edges, *W* is an *n x n* weight matrix and *r* is the memory retrieval rate calculated by the B-Matrix approach. In order to improve the retrieval rate of the network, the following algorithm is proposed:

Step 1: Identify the Inactive nodes, $\{n_1, n_2 ... n_i\}$ in a Network, where $i \leq n$
Step 2: Determine the memories $\{m_1, m_2 ... m_j\}$ to be retrieved
Step 3: Select a node $n_k$, where $1 \leq k \leq i$, to be triggered
Step 4: Apply the Widrow-Hoff Learning Rule to the weight matrix *W*
Step 5: Apply B-Matrix to the changed network
Step 6: If Retrieval rate > *r,* go to Step 2
     Else go to Step 3
Step 7: Repeat Step 6 until Retrieval rate =100 or Threshold = *t*

This algorithm gives a brief conception of the learning methodology. We will discuss each of the above steps in detail and will also show graphically the improved efficiency in the retrieval rates of different networks.

**Inactive Nodes**

We consider that the memories $\{m_1, m_2 ... m_l\}$ are stored in the network. We now calculate the T-Matrix and B-Matrix for the network. By clamping a small input fragment we retrieve memories and the memory vectors obtained from each of the *n* neurons is $\{v_1, v_2 ... v_n\}$. $\forall$ $v_k$ such that $1 \leq k \leq n$, compare it to each memory vector $m_j$ such that $1 \leq j \leq l$. If any of the vector $v_k$ is equal to either $m_j$ or - $m_j$ then that memory is said to be retrieved and the neuron that retrieves the memory is an active neuron. The neurons whose vectors are not same as any of the memories, such type of neurons are the *inactive* neurons.

Changes are made in the synapses of the inactive neurons such that they retrieve the desired memory when triggered by an input fragment. In some cases we do not have any inactive neurons, yet all the memories are not retrieved. In this case, the neuron that has generated the most frequently retrieved memory vector is selected as the "inactive" neuron.

**Memory to be retrieved**

If $\{m_1, m_2 ... m_l\}$ are the memories stored in the network, then on triggering the neurons with a small fragment, not all memory vectors are retrieved. The memory vectors that have not been retrieved are given as $\{m_1, m_2 ... m_j\}$ where $0 \leq j \leq l$. In order to select a memory that has to be retrieved at the chosen inactive node, we calculate the hamming distance between the vector that was originally retrieved at this node i.e. $v_k$ and $m_k$, where $1 \leq k \leq l$. The memory vector with the least hamming distance is chosen and if two or more have equal values, one of the memories is chosen randomly.

**Widrow-Hoff Learning Rule**

Once the inactive node and a suitable memory vector have been chosen, we apply the Widrow-Hoff learning rule to the B-Matrix. The B-Matrix is rearranged according to the selected nodes proximity. The error vector, B-Matrix and a threshold of *0.1* are the three parameters sent to the Widrow-Hoff function. To reduce the complexity of the process only the part of error vector and B-matrix in which the changes are to be made, is sent to the Widrow-Hoff function. This is a preeminent step by which the complexity of the process reduces significantly. The memories are retrieved from the modified B-Matrix to analyze the memory retrieval rate and also the transformation of inactive nodes to active nodes.

If the retrieval rate of the modified network is higher than the original network, the process is continued by choosing the next inactive node in order to retrieve more memories. Else, the modifications made to the weighted matrix are retraced and the memory with the penultimate hamming distance is chosen to be retrieved at that inactive node. This process continues till we have tried to retrieve all the memories at each inactive node or a retrieval rate of 100 percent is achieved.

## IV Results

The simulation was done in Matlab. The program takes into consideration the proximity of each neuron and calculates the memory retrieval using the simple B-Matrix approach. It determines all the inactive nodes and applies the learning algorithm to the network in order to retrieve the maximum number of memories. It also gives depicts the change between the active and inactive neurons. The following graphs show the results that have been obtained on different sized networks.

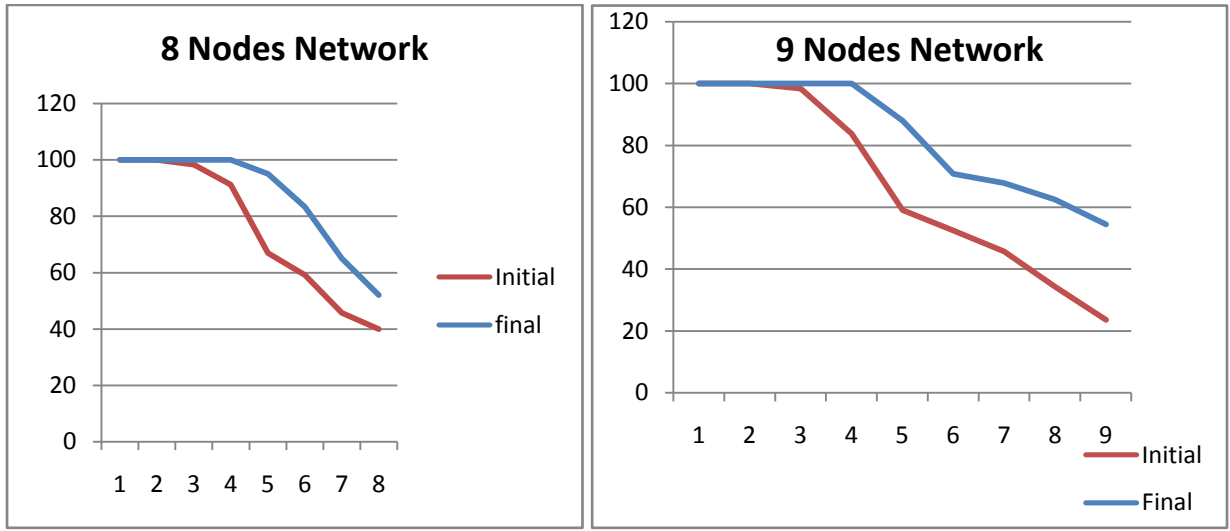

Figure 1: Graph depicting the retrieval rate before and after the learning algorithm in a 8 and 9 Node Networks.

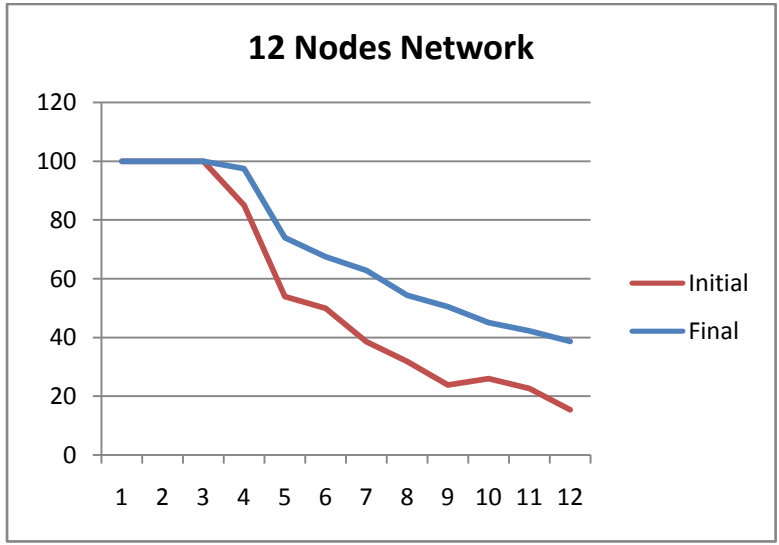

Figure 2: Graph depicting the retrieval rate before and after the learning algorithm in a 12 Node Network.

The following figure shows the number of active and inactive neurons in a network of 16 nodes which has been trained with 4 memories. Each color represents the each memory vector and the gray color indicates an inactive neuron.

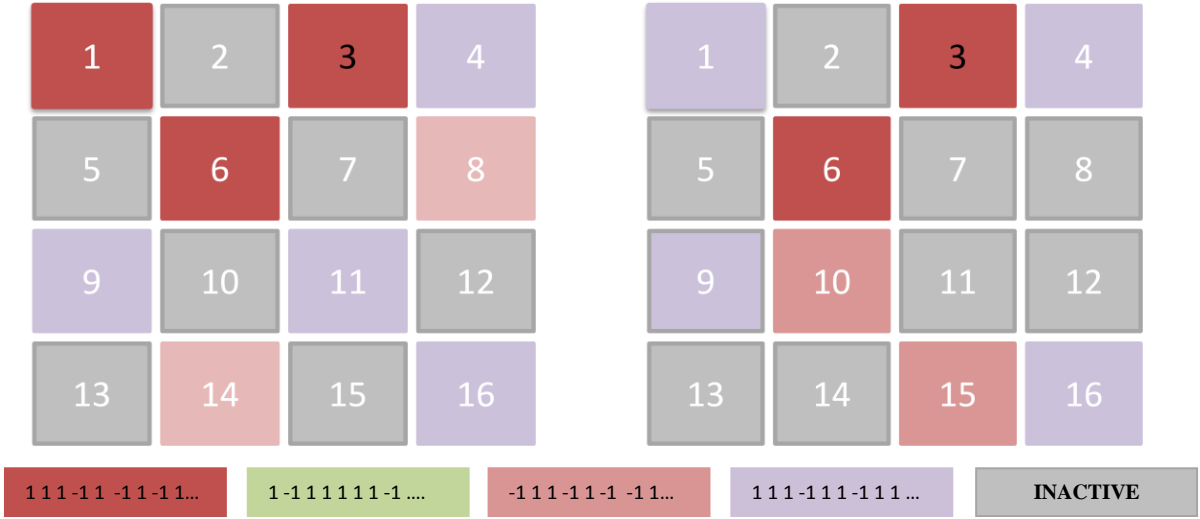

Figure 3a, 3b: Active and Inactive neurons when triggered with 1 and -1 respectively, before the learning algorithm

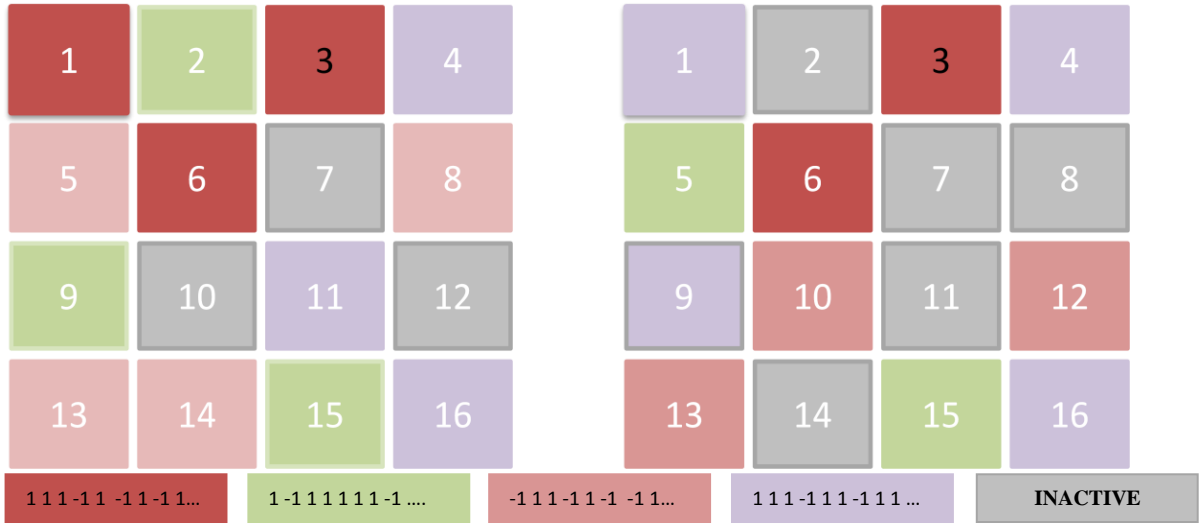

Figure 4a, 4b: Active and Inactive neurons when triggered with 1 and -1 respectively, before the learning algorithm

In the future work we will apply the learning algorithm to bigger networks and also analyze the activity of the inactive neurons in them to study the in depth properties of these neurons.

## Conclusion

In this paper, we investigated the use of delta learning to increase the memory retrieval rate from the B-Matrix. We also analyze the neurons that have become active after the learning algorithm is applied to the network. We have taken measures to reduce the computational complexity of the learning algorithm applied to the matrix.